\documentclass[letterpaper, 10 pt, conference]{ieeeconf}  

\IEEEoverridecommandlockouts                              

\usepackage{graphics} 
\usepackage{epsfig} 
\usepackage{mathptmx} 
\usepackage{times} 
\usepackage{amsmath} 
\usepackage{amssymb}  

\usepackage{amsfonts}
\usepackage{mathtools} 
\usepackage{amsmath} 
\usepackage{graphicx}
\usepackage{booktabs}
\usepackage{multirow}
\usepackage{colortbl}
\usepackage{svg}
\usepackage{wrapfig}
\usepackage{tabularx}
\usepackage{subfig}
\usepackage{makecell}
\usepackage{siunitx}
\usepackage{xcolor}  
\usepackage{float}
\usepackage[dvipsnames]{xcolor}

\usepackage{pifont}
\newcommand{\cmark}{\ding{51}}%
\newcommand{\xmark}{\ding{55}}%

\usepackage{hyperref}  
\hypersetup{
    colorlinks = true,
    linkcolor = red,
    anchorcolor = blue,
    citecolor = blue,
    filecolor = blue,
    urlcolor = blue,
    citebordercolor = red
    }

\title{\LARGE \bf
Autonomous Vehicle Controllers\\ From End-to-End Differentiable Simulation}

\author{
Asen Nachkov \quad Danda Pani Paudel \quad Luc Van Gool \\
INSAIT, Sofia University “St. Kliment Ohridski”, Sofia, Bulgaria
}

\begin{document}

\maketitle
\thispagestyle{empty}
\pagestyle{empty}

\begin{abstract}
  Current methods to learn controllers for autonomous vehicles (AVs) focus on behavioural cloning. Being trained only on exact historic data, the resulting agents often generalize poorly to novel scenarios. Simulators provide the opportunity to go beyond offline datasets, but they are still treated as complicated black boxes, only used to update the global simulation state. As a result, these RL algorithms are slow, sample-inefficient, and prior-agnostic. In this work, we leverage a differentiable simulator and design an analytic policy gradients (APG) approach to training AV controllers on the large-scale Waymo Open Motion Dataset. Our proposed framework brings the differentiable simulator into an end-to-end training loop, where gradients of the environment dynamics serve as a useful prior to help the agent learn a more grounded policy. We combine this setup with a recurrent architecture that can efficiently propagate temporal information across long simulated trajectories. This APG method allows us to learn robust, accurate, and fast policies, while only requiring widely-available expert trajectories, instead of scarce expert actions. We compare to behavioural cloning and find significant improvements in performance and robustness to noise in the dynamics, as well as overall more intuitive human-like handling. 
\end{abstract}


\section{Introduction}
\label{sec: intro}

When training AV controllers, it is common to treat the environment as a black-box function that is only used to evolve the states and provide an interactive element to the data collection process. Different RL algorithms adopt different strategies for handling it. Specifically,  value-based \cite{mnih2015human}, model-based \cite{silver2016mastering}, or policy gradient \cite{sutton1999policy} methods are oblivious to how the next state has been generated. They treat it as either a target, or some state from which to bootstrap a value estimate. This approach is general, but makes training slow and sample-inefficient, since the world dynamics and a behavioural policy have to be learned from the data. Imitation learning \cite{codevilla2018end}, being essentially supervised learning on expert actions, is more sample-efficient but still avoids utilizing the environment dynamics to help learn a policy.  

\begin{figure}[t]
    \centering
    \includegraphics[width=0.42\textwidth]{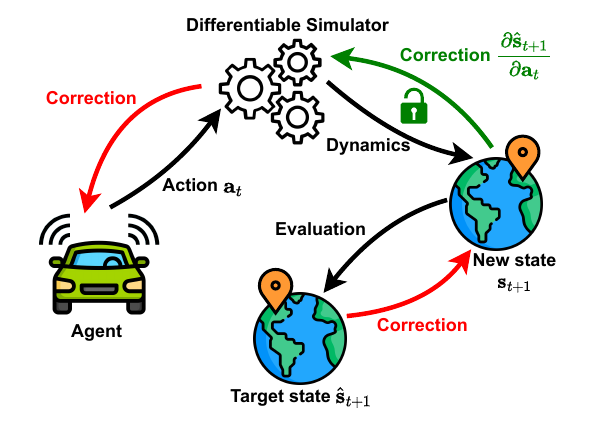}
    \captionsetup{belowskip=-0.3cm}
    \caption{\textbf{End-to-end learning of controllers.}
    Our framework uses \textcolor{OliveGreen}{\textbf{the gradients of the dynamics}} in a differentiable simulator to learn vehicle controllers from the corrections between the simulated new states and the target states.  
    }
    \label{fig: teaser}
\end{figure}

If the environment dynamics are known, one would likely be able to harvest the best aspects from each of these methods. In fact, in a differentiable environment one can optimize the policy directly using gradient descent, just via supervision from expert agent trajectories, as shown in Fig.~\ref{fig: teaser}. The benefits of this are: 1) Obtaining an \emph{explicit policy for continuous control}, 2) \emph{Fast inference}, since there is no planning at test time, 3) \emph{Unbounded policy precision}, due to the dynamics not being approximated, 4) Facilitating more \emph{grounded learning} by incorporating the dynamics directly into the training loop. This occurs from mixing the gradients of the dynamics with the usual derivatives, within the backpropagation, and using them to update the policy.

\begin{figure*}[h]
    \centering
    \includegraphics[width=1.0\textwidth]{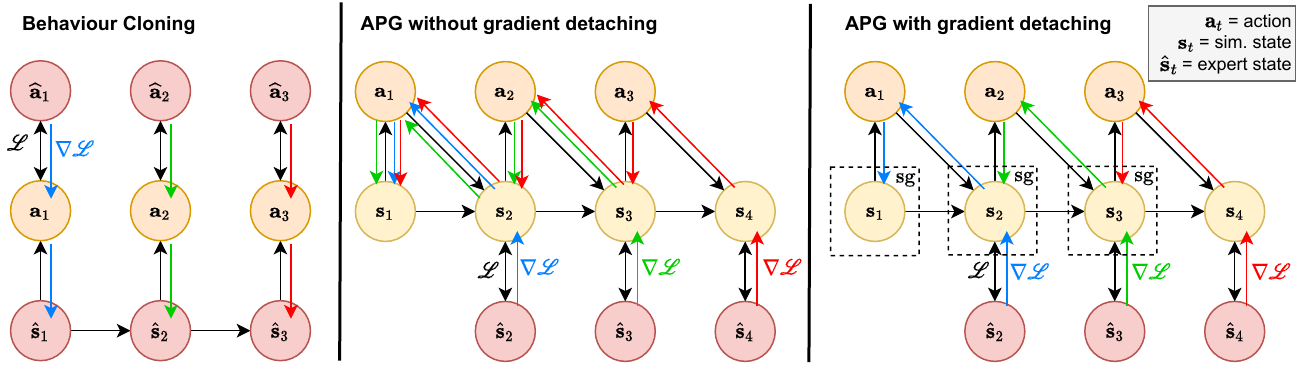}
    \captionsetup{belowskip=-0.3cm}
    \caption{\textbf{Learning with and without simulator.} Left: learning by behaviour cloning where we replay the GT trajectory and supervise the predicted actions. Middle: APG where we roll-out and supervise the trajectories without detaching gradients (shown in colored arrows). Right: APG where we detach gradients from past timesteps. The slanted arrows from $\textbf{a}_t$ to $\textbf{s}_{t+1}$ are the environment dynamics. The  proposed detachment during simulation offers efficient and lightweight training.}
    \label{fig: modeling_considerations}
\end{figure*}

\begin{figure*}[t]
    \centering
    \includegraphics[width=1.0\textwidth]{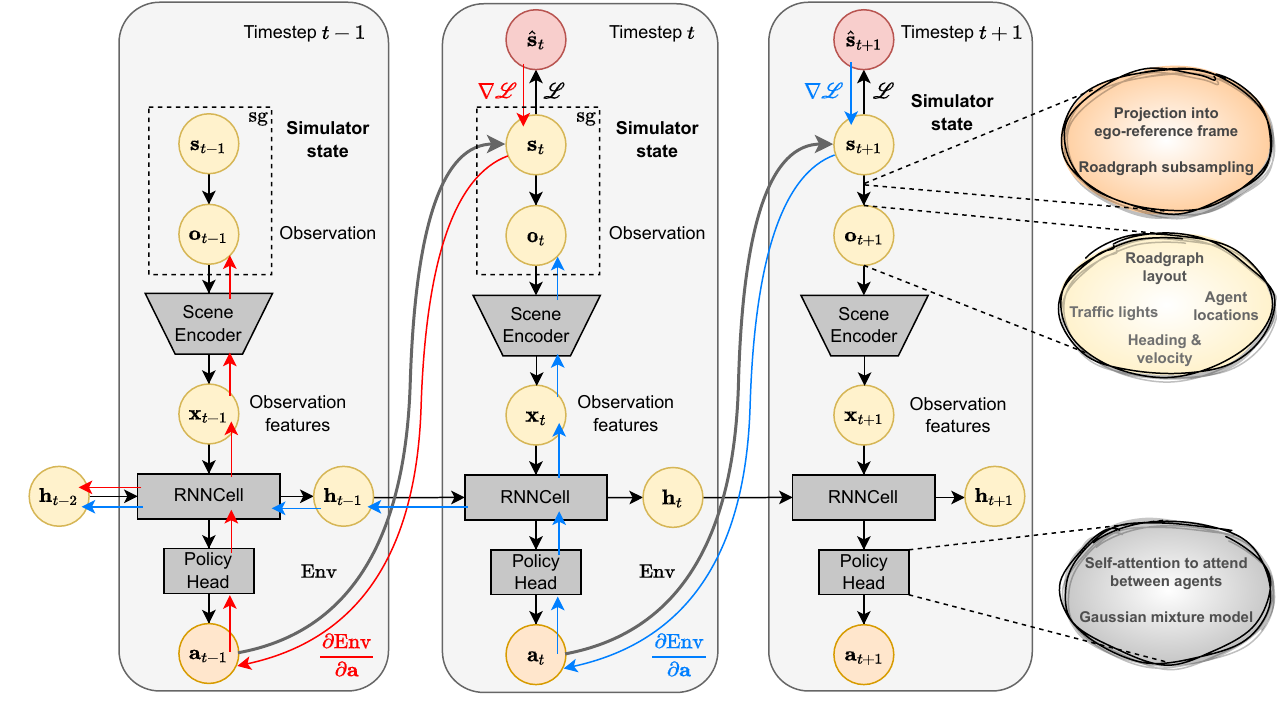}
    \captionsetup{belowskip=-0.3cm}
    \caption{\textbf{Unrolling the model in time with gradient detachment inside the differentiable simulator.} Starting from the simulator state $\mathbf{s}_t$, we obtain an observation $\mathbf{o}_t$, containing the scene elements such as agents locations, traffic lights, and roadgraph, which gets encoded into features $\mathbf{x}_t$. An RNN (recurrent over time) with a policy head outputs actions $\mathbf{a}_t$ which are executed in the simulated environment to obtain the new state $\mathbf{s}_{t + 1}$. When applying a loss between $\mathbf{s}_{t + 1}$ and $\hat{\mathbf{s}}_{t + 1}$ the gradients flow back through the environment and update the policy head, RNN, and the scene encoder. Similar to BPTT, gradients through the RNN hidden state accumulate. We do not backpropagate through the observation or the simulator state.}
    \label{fig: model_overview}
\end{figure*}

Waymax \cite{gulino2024waymax} was recently introduced as a differentiable, data-driven, large-scale autonomous driving simulator. In this work, we utilize its differentiability to train controllers using Analytic Policy Gradients (APG), reaping all the benefits mentioned above. Our proposed architecture is recurrent and propagates temporal information across long simulated trajectories, which allows the dynamics gradients to flow back to earlier steps (trace the blue arrows in Fig. \ref{fig: model_overview}), improving training speed and performance. To the best of our knowledge, we are the first to apply APG on such a large-scale task, obtaining strong competitive performance comparable to larger heavily-engineered AV models which do not make use of simulators.

\textbf{Contributions.} Our main contributions are:
\begin{enumerate}
    \item We identify and utilize a set of gradients obtainable within the Waymax simulator required for the downstream task of learning AV controllers, explained in Sec. \ref{subsec: diffthrough_waymax} In this process, we modify some operations in the dynamics to make them gradient-friendly.
    
    \item We develop a framework for end-to-end training of AV controllers using differentiable simulation and discuss its particular design implications and model constraints in Section \ref{subsec: training_setups} and \ref{subsec: model_overview}.
    
    \item We train and evaluate a model based on APG, in Sec.~\ref{sec: experiments}, which attains strong performance while being more accurate and more robust than relevant baselines.
\end{enumerate}

\section{Related Work}
\label{sec: related_work}

\textbf{APG}. Most RL methods that consider the environment as a black box (shown in Table \ref{table: related_work}) do not incorporate any world priors otherwise contained in the simulator. Compared to them, the analytic policy gradients approach (APG) is stable, sample efficient, and fast at test time. It has been used for trajectory tracking in drones \cite{wiedemann2023training}, including from visual input \cite{heeg2024learning}, quadruped control, \cite{song2024learning}, and is increasingly being used in differentiable physics simulators for robotic manipulation and navigation \cite{freeman2021brax, gillen2022leveraging, holl2020learning, mora2021pods, li2024visfly}.

\begin{table}[ht]
\centering
\begin{tabularx}{\columnwidth} { 
   >{\centering\arraybackslash}p{0.2\columnwidth}
   >{\centering\arraybackslash}p{0.4\columnwidth}
   >{\centering\arraybackslash}X  }
\toprule[1.5pt]
 \textbf{Method} & \textbf{Pros} & \textbf{Cons} \\
 \midrule[1pt]
 Value-learning \cite{mnih2015human, bellemare2017distributional, wang2016dueling, van2016deep}  & Stable, guaranteed convergence  & Implicit policy, moving target optimization  \\
 \midrule
 Policy gradients \cite{sutton1999policy, levine2016end, schulman2015trust, schulman2017proximal}  & Intuitive, scales well  & Unstable, sample-inefficient, converges to local optima \\
  \midrule
 Model-based \cite{silver2016mastering, schrittwieser2020mastering} & Offline, planning & Slow, bounded by precision of dynamics model  \\
  \midrule
 Imitation learning \cite{codevilla2018end, le2022survey}  & Scales well, offline  &  Non robust, generalization difficulty \cite{codevilla2019exploring}  \\
\bottomrule[1.5pt]
\end{tabularx}
\caption{\textbf{Different controller-training methods.} All of them consider the environment as a black box.}
\label{table: related_work}
\end{table}

\textbf{Relevance to BPTT}. APG is similar to backpropagation-through-time (BPTT) in that both typically differentiate the forward pass of a system component through multiple timesteps. In the case of RNNs \cite{hochreiter1997long, bahdanau2014neural} it is the application of a recurrent cell that is differentiated multiple times, whereas with APG it is the dynamics of an environment \cite{murthy2020gradsim}. Since in traditional sequence modeling (e.g. machine translation \cite{sutskever2014sequence}) the observed feature sequences do not depend on the recurrent model outputs, a better comparison to APG is optimizing recurrent policy networks \cite{wierstra2010recurrent, schaefer2007recurrent}. In such cases vanishing or exploding gradients have been observed, stemming from the spectrum of the Jacobian of the environment dynamics \cite{metz2021gradients}.

\textbf{Current approaches to trajectory prediction.} Recently, trajectory prediction for AVs has been dominated by transformer models \cite{shi2022motion, yuan2021agentformer, nayakanti2023wayformer, hu2023planning}. Specifically, MTR \cite{shi2022motion} uses a transformer encoder to predict an initial per-agent trajectory and collect map features alongside it. It then refines this trajectory using a transformer decoder. MVTA \cite{wang2023multiverse} adapts MTR for closed-loop simulation, as required by WOSAC \cite{montali2024waymo}, by introducing receding-horizon predictions and variable-length history aggregation. Other models focus on collision-avoidance \cite{chiu2023collision} or interactivity between agents in order to improve performance \cite{joint_multipath}. All these methods directly predict future locations, not actions, and therefore do not rely on any simulations.

\section{Method}
\label{sec: method}

In essence, a differentiable environment allows us to backpropagate gradients through it and directly optimize the policy. The resulting method broadly falls into the \emph{analytic policy gradients} (APG) type of algorithms. In our setting we assume we are given expert trajectories $ \{ \hat{\mathbf{s}}_t \}_{t=1}^T $, instead of rewards. The goal is to train the policy $\pi_\theta$ so that upon a roll-out, it reproduces the expert trajectory:
\begin{equation} \label{eq:task_objective}
\begin{aligned}
    \min_\theta \mathcal{L} &= \frac{1}{T} \sum_{t=1}^T  {\lVert \hat{\mathbf{s}}_t - {\mathbf{s}}_t \rVert}_2, \\ 
    \text{ where } {\textbf{s}}_t &= \text{Env}\left( {\textbf{s}}_{t-1}, \mathbf{a}_{t-1} \right) \text{ and } \textbf{a}_{t-1} \sim \pi_\theta \left( \textbf{s}_{t-1} \right)
\end{aligned}
\end{equation}
Here $\hat{\textbf{s}}_t$ and $\hat{\textbf{a}}_t$ refer to ground truth states and actions while $\textbf{s}_t$ and $\textbf{a}_t$ are the corresponding simulated states and actions. The sequence of states $\{\mathbf{s}_t \}_{t=1}^T$ forms a trajectory.

Optimization task \eqref{eq:task_objective} is difficult because trajectories are generated in a sequential manner with current states depending on previous actions, which themselves recursively depend on previous states. Additionally, $\nabla_\theta \mathcal{L}$ (we show for only one of its additive terms) has the form
\begin{equation} \label{eq: grad_form}
\frac{\partial \mathcal{L}}{\partial \theta} = \frac{\partial \mathcal{L}}{\partial {\mathbf{s}_t}}
\frac{\partial \mathbf{s}_t}{\partial {\mathbf{a}_{t-1}}}
\frac{\partial \mathbf{a}_{t-1}}{\partial {\mathbf{s}_{t-1}}}
\frac{\partial \mathbf{s}_{t-1}}{\partial {\mathbf{a}_{t-2}}}
\dots
\frac{\partial \mathbf{a}_0}{\partial \theta }
\end{equation}
which multiplicatively composes multiple derivatives $\frac{\partial \mathbf{s}_t}{\partial \mathbf{a}_{t-1}}$ corresponding to the environment dynamics themselves. Depending on the spectrum of the dynamics, these gradients may vanish or explode.

\textbf{Application setting.} Ultimately, for motion perception and planning in autonomous vehicles, we are interested in multi-modal trajectory prediction. Given a short history segment for all agents in the scene, we want to obtain $K$ \emph{modes}, or possible trajectories, for the future of each agent. In the presence of a simulator we can learn a stochastic policy, or controller, for the agents. Then, we can perform $K$ roll-outs, obtaining $K$ future trajectories. Hence, in the presence of a simulator, the task of trajectory prediction reduces to learning a policy $\pi_\theta: \textbf{s}_t \mapsto \textbf{a}_t$ to control the agents.

\subsection{Differentiating through Waymax} \label{subsec: diffthrough_waymax}

\textbf{Obtaining gradients.} We apply our APG method in the Waymax simulator \cite{gulino2024waymax}. Being implemented in Jax \cite{jax2018github}, it is relatively flexible in choosing which variable to differentiate and with respect to what. However, one is now faced with the task of choosing which gradients to use for training and how. Some are unavailable or not meaningful. For example,
\begin{enumerate}
    \item The derivatives of future agent locations with respect to current traffic light states or roadgraph points are all zero, because the simulator dynamics (e.g., bicycle or delta dynamics) do not depend on the roadgraph or the traffic lights. 
    
    \item Certain metrics such as collision or offroad detection are boolean in nature. Other objects such as traffic lights have discrete states. While useful for training, these are problematic for differentiation.
\end{enumerate}

What is \emph{meaningful} and \emph{useful} is to take the gradients of future controllable agent locations with respect to their current actions, $\partial \mathbf{s}_t / \partial \mathbf{a}_{t-1}$. These are precisely the derivatives needed for objective \eqref{eq:task_objective} and hence we focus on them. Since the simulator state is represented as a set of tensors, and the motion dynamics are computed as continuous functions over them, we treat the dynamics simply as another operation in the computation graph, and apply Jax's tracing, just-in-time compilation, and functional transformations like \texttt{grad} over it without much effort.

\begin{figure}[t]
    \centering
    \includegraphics[width=1.00\columnwidth]{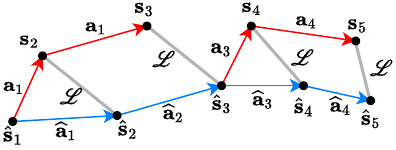} 
    \captionsetup{aboveskip=0cm, belowskip=-0.5cm}
    \caption{\textbf{ Resetting agent's state during incremental learning.} Once every $n$ steps, we reset the agent's position to the corresponding log state. \textcolor{NavyBlue}{Blue} arrows show the GT trajectory, \textcolor{red}{red} is the discontinuous simulated trajectory, and \textcolor{gray}{\textbf{gray}} lines show the supervision.}%
    \label{fig:incr_learning}%
\end{figure}

Moreover, we found it useful to adapt the bicycle dynamics model to be gradient-friendly. This includes adding a small epsilon to the argument of a square root to avoid the case when its input is 0, as well as adapting the yaw angle wrapping, present in many similar settings, to use \texttt{arctan2} instead of \texttt{mod}, which makes the corresponding derivatives continuous and facilitates training.

\begin{figure*}[h]
    \centering
    \includegraphics[width=1\textwidth]{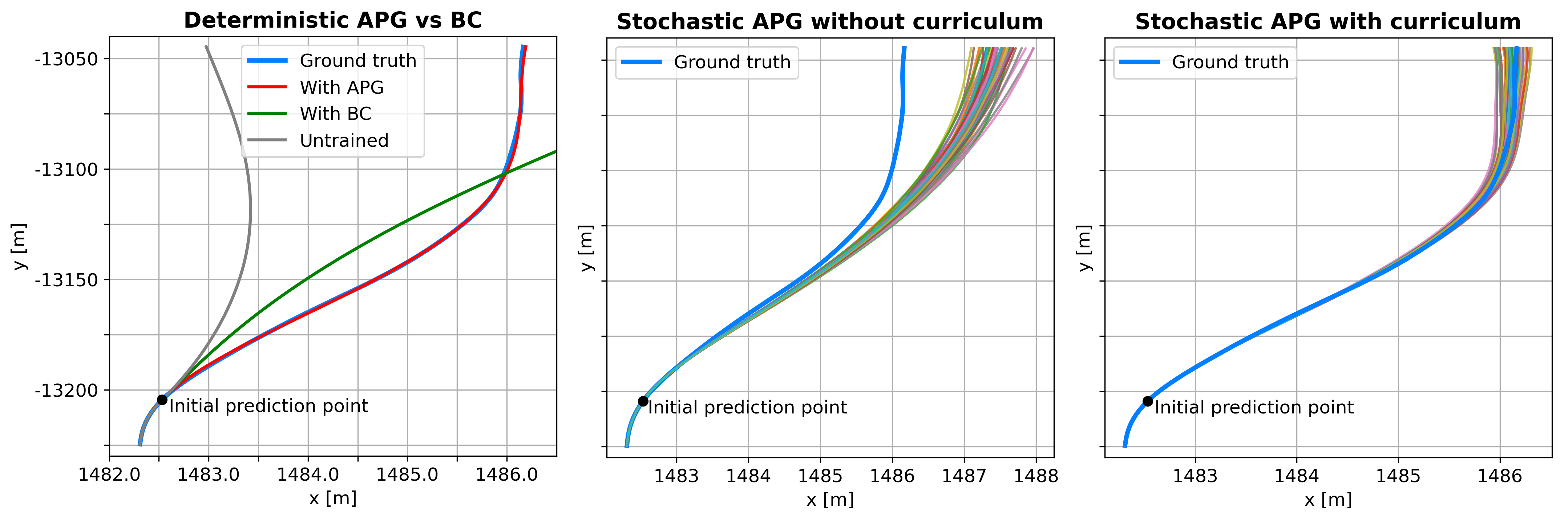}
    \captionsetup{aboveskip=0cm, belowskip=-0.1cm}
    \caption{\textbf{Initial results on a toy task.} $x$ and $y$ axes represent spatial coordinates. Left: deterministic beh. cloning fails to reproduce the trajectory on which it was fitted, while APG succeeds. Middle: APG struggles with a stochastic policy because of the sequential nature of the task. Right: APG with periodical resetting while training improves sample efficiency.}
    \label{fig: toy_task1}
\end{figure*}

\subsection{End-to-end training with the simulator} \label{subsec: training_setups}
\textbf{Dense trajectory supervision.} Obtaining the gradients of the environment dynamics opens up technical questions of how to train in such a setup. One can supervise the rolled-out trajectory only at the final state and let the gradients flow all the way back to the initial state. Since this does not supervise the \emph{path} taken to the objective, in our experiments we densely supervise all states in the collected trajectory with all states in the GT trajectory.

\textbf{Gradient detaching.} 
Dense supervision allows us to detach the gradients at the right places, as shown in the third part of Fig. \ref{fig: modeling_considerations}. Here, when we obtain $\mathbf{s}_t$, we calculate the loss and backpropagate through the environment dynamics obtaining $\partial \mathbf{s}_t / \partial \mathbf{a}_{t - 1}$ without continuing on to previous steps. This makes training slower, since gradients for the earlier steps do not accumulate, as in a RNN, but effectively cuts the whole trajectory into many $(\textbf{s}_t, \textbf{a}_t, \textbf{s}_{t+1})$ transitions which can be treated independently by the optimization step. This allows for \emph{off-policy} training -- a key aspect of our setup.

\subsection{Model overview} \label{subsec: model_overview}
We present our model setup in Figure \ref{fig: model_overview}. For training, we roll-out a full trajectory and supervise with the GT one. The gradients flow back through the differentiable dynamics, policy and scene encoder, and continue back to the previous scene contexts using a RNN hidden state. We detach the previous simulator state for both \emph{necessity} and \emph{flexibility} so that the state transitions on which we compute the loss can be different from the transitions executed during roll-out.

\textbf{Optimization difficulty.} Differentiating through a stochastic policy requires that the actions be reparametrized \cite{kingma2013auto}. But every reparametrization leads to stochastic gradients for those layers before it. And when we compose sampling operations sequentially, such as in the computation graph of the entire collected trajectory, the noise starts to compound and may overwhelm the actual signal from the trajectory steps, making the optimization more noisy and difficult. To address this, we implement \emph{incremental} training where we periodically ``reset" the simulated state back to the corresponding log state \cite{wiedemann2023training}, shown in Fig. \ref{fig:incr_learning}. This ensures that the data-collection stays around the GT trajectory, instead of far from it, increasing sample efficiency.
For incremental learning, we increase the difficulty by reducing the reset frequency as  training progresses.

\textbf{Agent processing.} For the multi-agent experiments described in Sec. \ref{sec: womd_experiments} we slightly adapt our architecture to allow training with $N$ agents but evaluating with $M$. We adopt a small transformer \cite{vaswani2017attention} that forces each agent to attend to the locations of the other agents in parallel. However, at the last transformer block we use a fixed number of learned queries which attend over the variable number of agent keys and values, effectively soft-clustering them, and becoming independent of their number. This allows us to train with 32 agents but evaluate with 128, as required by \cite{ettinger2021large}.

\section{Experiments}
\label{sec: experiments}

\begin{table*}[t]
    \small
    \centering
    \begin{tabular}[width=\textwidth]{ 
        p{0.14\textwidth} | p{0.13\textwidth} |  p{0.09\textwidth} | p{0.04\textwidth} | p{0.09\textwidth} || S[table-format=2.4] | S[table-format=2.4] | S[table-format=2.4] 
    } 
         \toprule[1.5pt]
         \rowcolor[gray]{0.9}
         \textbf{Method} & \textbf{Task} & \textbf{Dynamics} & \textbf{IDM} & \textbf{Supervise} & 
         \multicolumn{1}{c}{\makecell{min \\ ADE $\downarrow$}} & 
         \multicolumn{1}{c}{\makecell{min \\ overlap $\downarrow$}} & 
         \multicolumn{1}{c}{\makecell{min \\ offroad $\downarrow$}} \\
         \midrule[1pt]
         Rand. action & Plan & Delta & -- & -- &  16.3925 & 0.5292 & 0.4989 \\
         Const. velocity & Plan & Delta & -- &  -- &  5.9547 & 0.3172 & 0.1150 \\
         Beh. cloning & Plan & Bicycle & \xmark & Actions &  \textbf{1.4157} & 0.2475 & 0.2673 \\
         Beh. cloning & Plan & Bicycle & \cmark & Actions &  1.4868 & 0.2647 & 0.2132 \\
         Beh. cloning & Plan & Delta & \xmark & Actions & 3.0063  & 0.1741  & 0.0278  \\
         Beh. cloning & Plan & Delta & \cmark & Actions & 3.2456 & 0.1910 & 0.0330 \\
         \rowcolor[gray]{0.95}
         APG, stoc.  & Plan & Bicycle & \xmark & States &  2.0083 & \textbf{0.0800} & 0.0282 \\
         \rowcolor[gray]{0.95}
         APG, stoc  & Plan & Bicycle & \cmark & States &  6.0201 & 0.2710 & 0.0560 \\
         \rowcolor[gray]{0.95}
         APG, stoc. & Plan & Delta & \xmark &  States &  2.0793 & 0.0870 & \textbf{0.0200} \\
         \rowcolor[gray]{0.95}
         APG, stoc. & Plan & Delta & \cmark & States &  3.7624 & 0.2120 & 0.0370 \\
         \midrule[1pt]
         Wayformer & Plan & Delta & \cmark & States & 2.3800 & 0.1068 & 0.0789 \\
         Beh. cloning & Plan & Delta & \cmark & Actions & 6.2800 & 0.0583 & 0.0414 \\
         Beh. cloning & Plan & Bicycle & \cmark & Actions & 3.6000 & 0.1120 & 0.1359 \\
         Beh. cloning & Plan & Bicycle* & \cmark & Actions & 2.2600 & 0.0459 & 0.0110 \\
         Beh. cloning & Plan & Delta* & \cmark & Actions & 2.9800 & 0.0597 & 0.0442 \\
         DQN & Plan & Bicycle* & \cmark & Rewards & 9.8300 & 0.0650 & 0.0374 \\
         DQN & Plan & Bicycle* & \cmark & Rewards & 10.7400 & 0.0491 & 0.0431 \\
         \midrule[1pt]
         Rand. action & Multi-agent & Bicycle & -- & -- &  29.3291 & 0.8276 & 0.8531 \\
         Const. velocity & Multi-agent & Bicycle & -- & --  &  9.2058 & 0.6997 & 0.7105 \\
         Beh. cloning & Multi-agent & Bicycle & --  & Actions &  4.1123 & 0.8527 & 0.999 \\
         APG & Multi-agent & Bicycle &  -- & States &  1.8096 & 0.6021 & 0.9887 \\
         Beh. cloning & Multi-agent \textdagger & Bicycle & --  & Actions &  9.6331 & 0.7587 & 0.9508 \\
         APG & Multi-agent \textdagger & Bicycle &  -- & States &  3.7169 & 0.3953 & 0.5851 \\
         \bottomrule[1.5pt]
    \end{tabular}
    \captionsetup{belowskip=-0.1cm}
    \caption{\textbf{Experimental results.} We evaluate APG and other baselines on the WOMD \texttt{val} set. The results in the middle block are taken from \cite{gulino2024waymax} as a comparable reference and all include route conditioning. * refers to the discrete version of the action space. Behavioural cloning and APG for planning have been evaluated with 32 modes. \textdagger  refers to evaluating only on those objects for which \texttt{is\_modeled} is set by the WOM dataset.}
    \label{table: waymax_results}
\end{table*}

\begin{table*}[h]
    \small
    \centering
    \begin{tabular}[width=\textwidth]{ p{0.14\textwidth} | p{0.06\textwidth} |  p{0.09\textwidth} | p{0.04\textwidth} | p{0.14\textwidth} || >{\raggedleft\arraybackslash}p{0.09\textwidth} | >{\raggedleft\arraybackslash}p{0.09\textwidth} | >{\raggedleft\arraybackslash}p{0.09\textwidth} } \toprule[1.5pt]
         \rowcolor[gray]{0.9}
         \textbf{Method} & \textbf{Task} & \textbf{Dynamics} & \textbf{IDM} & \textbf{Supervise} & \makecell{min \\ ADE $\downarrow$} & \makecell{min \\ overlap $\downarrow$} & \makecell{min \\ offroad $\downarrow$} \\
         \midrule[1pt]
         APG, stoc. & Plan & Delta & \xmark &  $(x, y)$ &  2.0806 & 0.5671 & 0.3634 \\
         APG, stoc. & Plan & Delta & \cmark &  $(x, y)$ &  2.9700 & 0.5350 & 0.3863 \\
         APG, stoc. & Plan & Delta & \xmark & $(x, y, v_x, v_y, \theta)$ &  \textbf{2.0473} & \textbf{0.0880} & \textbf{0.0200} \\
         APG, stoc. & Plan & Delta & \cmark &  $(x, y, v_x, v_y, \theta)$ &  3.7624 & 0.2120 & 0.0376 \\
         \bottomrule[1.5pt]
    \end{tabular}
    \caption{\textbf{Ablation on supervising the velocity and yaw angle.} The ADE is evaluated only on $(x, y)$ locations. We found that if we supervise not only these locations, but also the velocity and yaw angle of the agents, results are better, particularly in terms of reduced collision and offroad rates.}
    \label{table: ablation_yaw_supervision}
\end{table*}

\begin{table*}[h]
    \small
    \centering
    \begin{tabular}[width=\textwidth]{ p{0.12\textwidth} || >{\raggedleft\arraybackslash}p{0.12\textwidth} | >{\raggedleft\arraybackslash}p{0.12\textwidth} | >{\raggedleft\arraybackslash}p{0.14\textwidth} | >{\raggedleft\arraybackslash}p{0.12\textwidth} | >{\raggedleft\arraybackslash}p{0.14\textwidth} } \toprule[1.5pt]
         \rowcolor[gray]{0.9}
         \textbf{Method} & ADE $\downarrow$ & overlap $\downarrow$ & overlap perc. $\downarrow$ & offroad $\downarrow$ & offroad perc. $\downarrow$ \\
         \midrule[1pt]
         BC-MA & 4.1123 & 0.8527 & 0.0590 & 0.9990 & 0.3657 \\
         APG-MA & \textbf{1.8096} & \textbf{0.6021} & \textbf{0.0319} & \textbf{0.9887} & \textbf{0.2983} \\
         BC-MA \textdagger & 9.6331 & 0.7587 & 0.0932 & 0.9508 & 0.3657 \\
         APG-MA \textdagger & \textbf{3.7169} & \textbf{0.3953} & \textbf{0.0364} & \textbf{0.5851} & \textbf{0.2983} \\
         \bottomrule[1.5pt]
    \end{tabular}
    \captionsetup{aboveskip=0.1cm, belowskip=-0.3cm}
    \caption{\textbf{Detailed metrics for multi-agent settings.} We compare beh. cloning and APG on multi-agent simulation with bicycle dynamics. \textdagger  refers to evaluating only on those objects for which \texttt{is\_modeled} is set by the WOM dataset. In the perc. columns we compute the metric by averaging over all individual agent transitions, not over entire agent trajectories.}
    \label{table: detailed_comparison_ma}
\end{table*}

\textbf{An early example.} To demonstrate the characteristics of APG, we first compare APG and behaviour cloning (BC) on a toy task -- purposefully overfitting a single trajectory for a single agent. We are given a full-length ground-truth (GT) trajectory from the WOM dataset \cite{ettinger2021large}, $\{ \hat{\mathbf{s}}_t \}_{t = 1}^T \in \mathbb{R}^{T \times 2}$, and are learning a controller $\pi_\theta$ of the form $\pi_\theta: (\textbf{s}_t, t) \mapsto \textbf{a}_t$.

\textbf{Behaviour cloning baseline.} Since expert actions are available only on the GT trajectory, behaviour cloning replays it and is reduced to supervised learning on them: $
\min_\theta \frac{1}{T} \sum_{t=1}^T {\lVert \hat{\mathbf{a}}_t - {\mathbf{a}}_t \rVert}_2$, where ${\textbf{a}}_t = \pi_\theta \left( \hat{\textbf{s}}_{t-1} \right)$. In that setup at test time, when we perform a roll-out, behaviour cloning suffers from error-compounding and is unable to reproduce the GT trajectory, even though it has learned the on-trajectory actions very accurately. Very small imprecisions in the actions start to accumulate, as shown in the left part of Fig.~\ref{fig: toy_task1}.

\textbf{APG}. If we have access to a simulator we can perform roll-outs at train time and supervise the resulting trajectories $\{ \textbf{s}_t \}_{t=1}^T$ with $ \{  \hat{\textbf{s}}_t \}_{t=1}^T$. This is sufficient to make the controller robust and allows it to reproduce the trajectory at test time on its own. If the policy is stochastic, the learning process becomes less sample-efficient because it becomes exponentially less likely that in a far away $t$, a given location $(x, y)$ is reached. Thus, as shown in Figure \ref{fig: toy_task1} converging to the GT trajectory does not happen uniformly from all directions, but rather sequentially. To improve sample-efficiency, as a form of \emph{curriculum}, we snap back the simulated trajectory whenever it goes beyond a threshold $\xi$ with respect to $\{  \hat{\mathbf{s}}_t \}_{t=1}^T$. In Fig \ref{fig:incr_learning} the resetting is based on time (once every $n$ steps), but here we reset based on distance. We also detach the gradients as shown previously in Fig. \ref{fig: modeling_considerations}.

Thus, in the extreme case when there are no additional regressors such as the roadgraph, or past agent locations, behaviour cloning fails in reproducing a GT trajectory on its own while APG does not.

\subsection{Large-scale experiments in Waymax}
\label{sec: womd_experiments}
\textbf{Experiment setup.} The Waymax simulator uses the underlying Waymo Open Motion Dataset (WOMD) \cite{ettinger2021large} to initialize the simulator state. It contains close to 500K training scenarios, each of length 9 seconds and a framerate of 10 Hz. The history length is 10 frames (1 second) and the goal is to predict the next 80 frames. To train the model, we run multiple simulations, one per scenario, in parallel (this being the simulated \emph{minibatch}) and use Adam to optimize them. We iterate through all scenarios (the equivalent of an \emph{epoch}) multiple times. At test time, we simulate multiple roll-outs and compute the average displacement error (ADE), offroad rate, and collision (also called overlap) rate for each mode. A collision rate of 20\% means that 20\% of all trajectories have \emph{some} objects colliding at some timestep in them. Since the future is uncertain and we want to make sure that the GT locations are covered by at least one of our modes, we report the eval metrics from the best mode. Thus, we evaluate on minADE, min offroad rate, and min collision rate.

\begin{figure}[!tbp]
    \centering
    \includegraphics[width=0.5\textwidth]{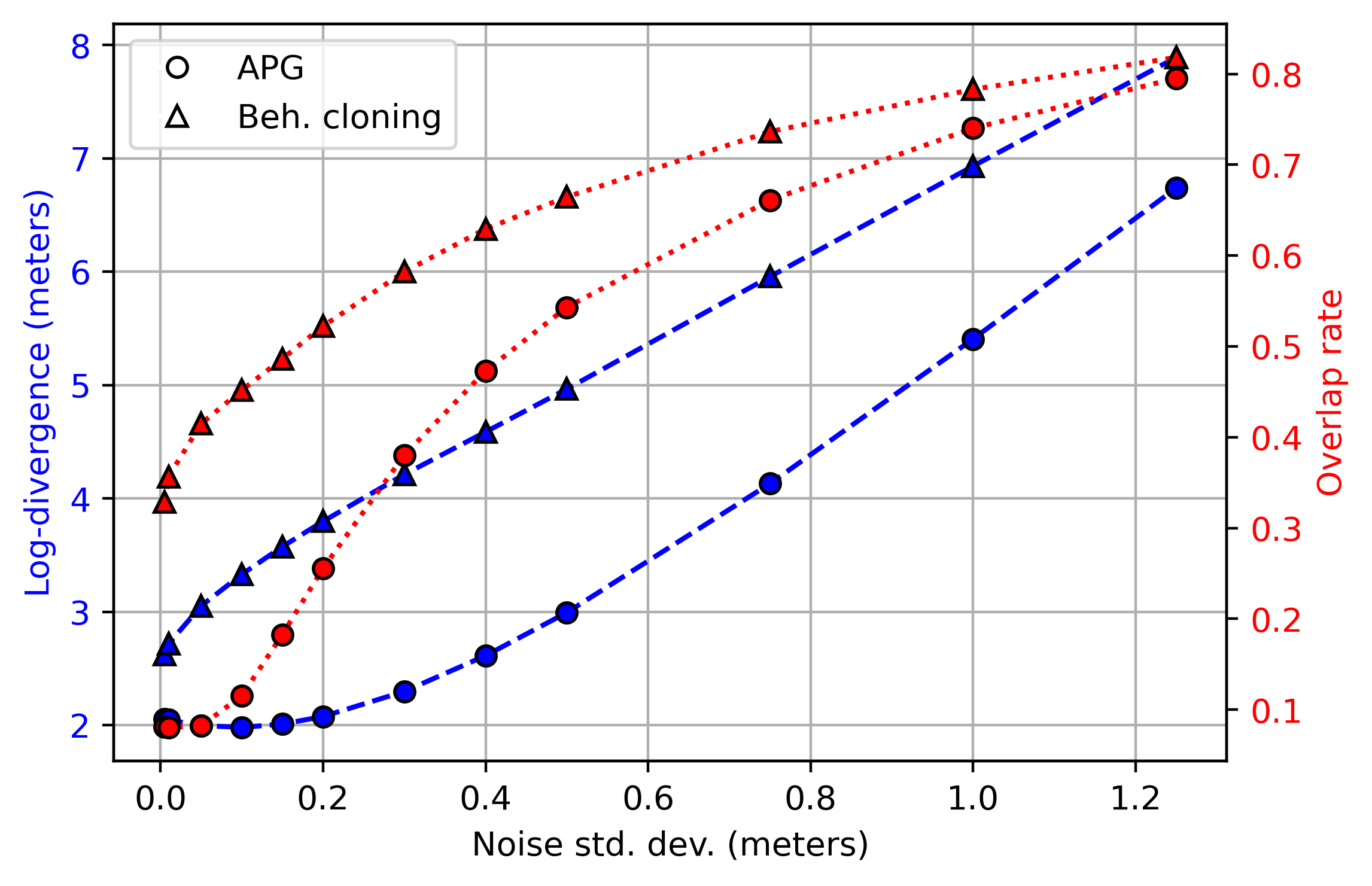}
    \caption{\textbf{Performance comparison in a stochastic environment.} Compared to BC, the performance of APG under increasingly noisy dynamics deteriorates later. For simplicity, we do not show the offroad rates, which are similar to the overlap ones for both APG and BC.}
    \label{fig: noisy_env}
    \vspace{-0.4cm}
\end{figure}

\begin{figure}[b]
    \centering
    \includegraphics[width=0.5\textwidth]{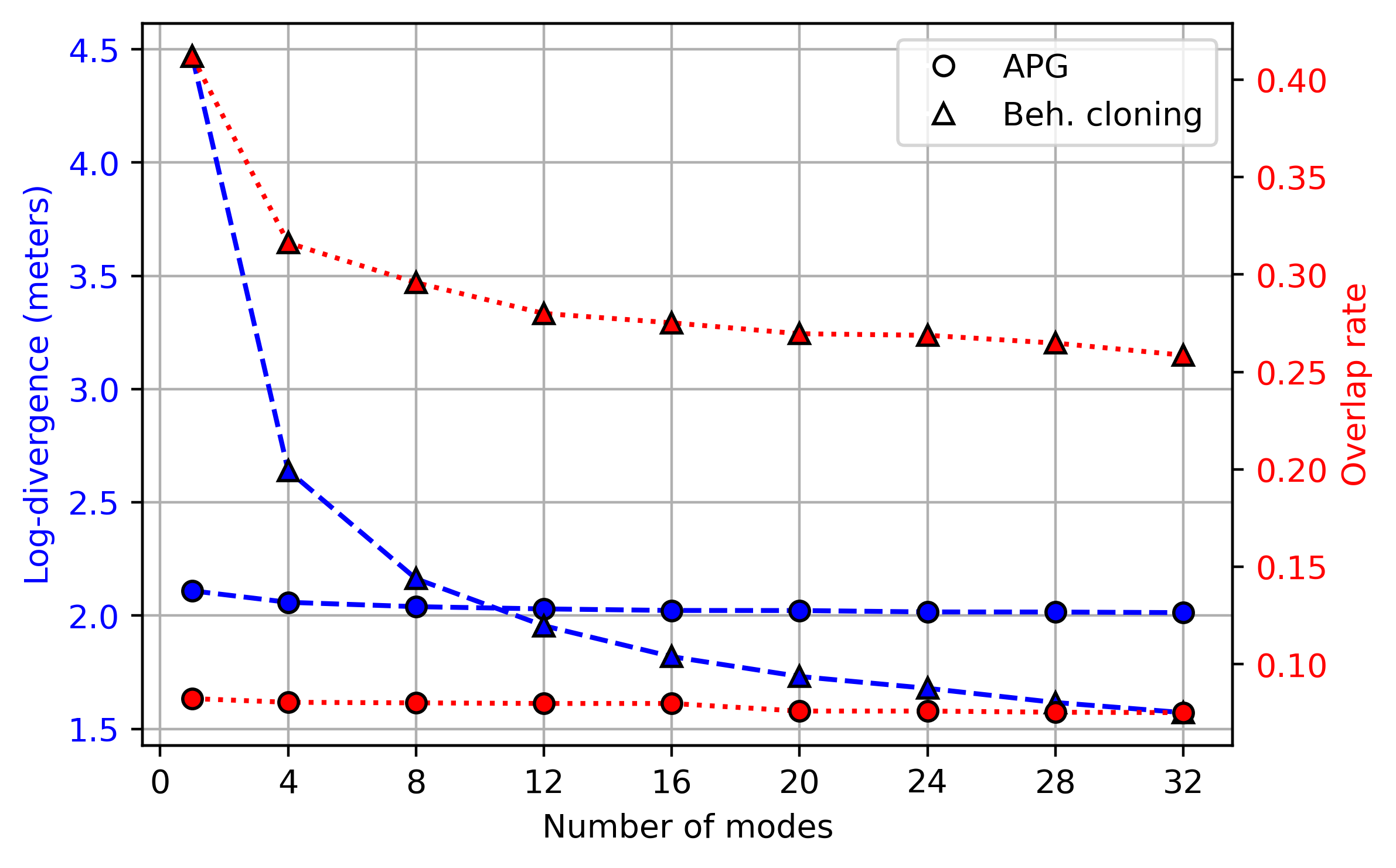}
    \caption{\textbf{Effect of the number of modes.} Compared to BC (triangles), APG (circles) is more confident and the performance is less dependent on the number of modes. }
    \label{fig: num_modes}
\end{figure}

\begin{figure*}[t]
    \centering
    \includegraphics[width=0.99\textwidth]{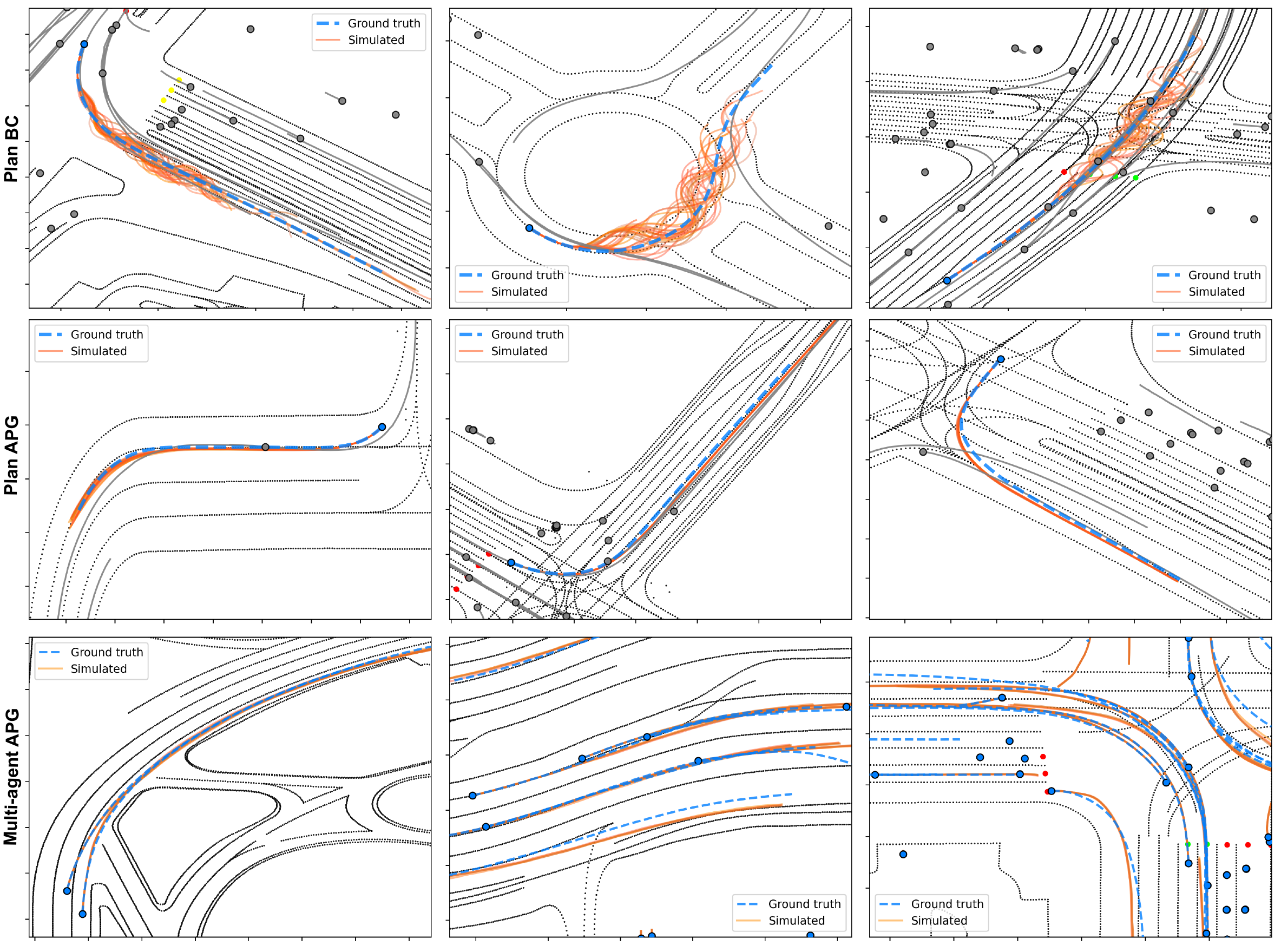}
    \captionsetup{aboveskip=-0.1cm, belowskip=-0.55cm}
    \caption{\textbf{Sample trajectories.} Row 1 shows BC, while row 2 and 3 show APG applied in the planning and multi-agent control. BC produces agents that swerve, while the APG trajectories are more realistic. Blue circles show controlled agents in the beginning of the trajectory. Red, green, and yellow circles represent traffic lights.}
    \label{fig: samples}
\end{figure*}

\textbf{Baseline}. We highlight that to the best of our knowledge, we are the first to apply APG in the Waymax + WOMD setting \cite{gulino2024waymax, ettinger2021large}. Hence, as the main baseline we set a non-APG method based on behavioural cloning -- the currently dominant approach to train AV policies. 

\textbf{Evaluation settings.} Our main results are shown in Table \ref{table: waymax_results}. We compare behavioural cloning (BC) with APG on two tasks -- \emph{planning}, where we learn a controller only for the self-driving car, and \emph{multi-agent} (MA), where we learn a single controller for all agents. For the latter setting, we project the roadgraph and agent coordinates in each agent's own coordinate frame. Hence, our model is \emph{agent}-centric. Agents' actions are selected independently, where each agent considers the location of all others. Multi-agency results from all of them being \emph{controlled} by the trained policy. 

\textbf{Training and evaluation specifics.} For the planning task we evaluate with and without IDM \cite{gulino2024waymax}, which makes other agents reactive to the ego-vehicle, instead of simply following the log trajectories. For MA-BC the training loss is applied over all simulated agents which have a valid GT action. For MA-APG it is applied over all agent transitions for which both the current sim agent and log agent are considered valid by the simulator. The general principle is that we must supervise the transitions of all agents controlled by the policy. We evaluate MA in two settings -- on all valid objects, and on all objects for which \texttt{is\_modeled} is set (the objects to predict for) as designated by WOMD \cite{montali2024waymo}.

\textbf{Input features.} The inputs to the model consist of the current agent locations, nearest roadgraph points, current traffic light states, and a heading direction to condition the agent roughly in terms of where it should go. The same conditioning is used in the BC baseline.

\textbf{Results analysis.} Our APG method outperforms those based on behavioural cloning (BC), obtaining better ADE, collision and offroad rates. In a few settings BC obtains a lower min ADE value, but qualitative results suggest that the agent swerves, producing unrealistic behaviour in terms of collision and offroad rates. The corresponding APG trajectories are considerably more realistic. Without IDM agents the testing setting is maximally similar to the training setting and APG performs best, with considerable improvement in overlap rate compared to BC. A design limitation of APG is that adding IDM during evaluation causes the learned motion dynamics to be different from the ones seen at test time, which naturally hurts performance. We view this as expected, given its supervised trajectory training and leave it as future work to perform additional finetuning with IDM agents.

APG is also more certain in terms of which actions to take, leading to lower variability between individual trajectory modes. Fig. \ref{fig: num_modes} shows that APG requires many fewer modes in order to reach a given performance. Thus, by producing more confident trajectories, it allows us to save on computation.

\textbf{Supervision of yaw angle and velocity.} Even though the log divergence metric only considers $(x, y)$ coordinates, we supervise also with the yaw and $x$-, $y$- velocities. Table \ref{table: ablation_yaw_supervision} shows that this is very beneficial for performance.

\textbf{Robustness to noise and trajectory length.} To assess potential sim-to-real transfer, we make the environment dynamics stochastic at test time by adding different levels of noise to the next agent locations. Fig. \ref{fig: noisy_env} shows that our APG method compares favourably to BC. For very small noise levels the performance is not affected at all. Additionally, in Table \ref{table: half_seq} we show that training on half of each sequence, instead of the full one, reduces performance only marginally.

\begin{table}[b]
    \vspace{-0.2cm}
    \small
    \centering
    \begin{tabular}[width=\textwidth]{ p{0.36\columnwidth} || >{\raggedleft\arraybackslash}p{0.12\columnwidth} | >{\raggedleft\arraybackslash}p{0.14\columnwidth} | >{\raggedleft\arraybackslash}p{0.14\columnwidth} } \toprule[1.5pt]
         \rowcolor[gray]{0.9}
         \textbf{APG, plan} & ADE $\downarrow$ & overlap $\downarrow$ &  offroad $\downarrow$  \\
         \midrule[1pt]
         Full trajectory & 2.0083 &  0.0800 & 0.0282 \\
         Half trajectory & 2.0493 & 0.0882 & 0.0256\\
         \bottomrule[1.5pt]
    \end{tabular}
    \captionsetup{aboveskip=0.2cm, belowskip=-0.4cm}
    \caption{\textbf{Shorter training sequences.} The performance is robust even when training on half the sequence lengths.}
    \label{table: half_seq}
\end{table}

\textbf{Qualitative study.} We showcase samples from our method in Fig. \ref{fig: samples} and observe that in the planning setting APG realizes accurate stochastic trajectories. Errors occur mostly from not predicting the correct acceleration, not the steering, which is prevalent in the BC models and causes the agent there to swerve. The MA-APG trajectories are relatively less accurate because, unlike in planning, when all agents are controlled, any prediction will have an effect on any agent, making training more difficult. Yet the APG performance is still better than the BC one, shown in Table \ref{table: detailed_comparison_ma}.
\section{Conclusion}
\label{sec: conclusion}

In this work, we leveraged the differentiable nature of Waymax to train AV policies for planning and multi-agent control. We proposed measures -- incremental learning, and a specific architecture based on an RNN for the temporal information transfer, and a transformer, for the spatial mixing of agent features -- to efficiently train long episodes in diverse settings. Results suggest that our model is accurate and robust, while at the same time being lightweight and fast at inference time, enabling potential real-world usage.

\textbf{Acknowledgements}. This research was partially funded by the Ministry of Education and Science of Bulgaria (support for INSAIT, part of the Bulgarian National Roadmap for Research Infrastructure).


\bibliography{citations}

\begin{thebibliography}{10}

\bibitem{bahdanau2014neural}
Dzmitry Bahdanau, Kyunghyun Cho, and Yoshua Bengio.
\newblock Neural machine translation by jointly learning to align and translate.
\newblock {\em arXiv:1409.0473}, 2014.

\bibitem{bellemare2017distributional}
Marc~G Bellemare, Will Dabney, and R{\'e}mi Munos.
\newblock A distributional perspective on reinforcement learning.
\newblock In {\em International conference on machine learning}, pages 449--458. PMLR, 2017.

\bibitem{jax2018github}
James Bradbury, Roy Frostig, Peter Hawkins, Matthew~James Johnson, Chris Leary, Dougal Maclaurin, George Necula, Adam Paszke, Jake Vander{P}las, Skye Wanderman-{M}ilne, and Qiao Zhang.
\newblock {JAX}: composable transformations of {P}ython+{N}um{P}y programs, 2018.

\bibitem{chiu2023collision}
Hsu-kuang Chiu and Stephen~F Smith.
\newblock Collision avoidance detour for multi-agent trajectory forecasting.
\newblock {\em arXiv:2306.11638}, 2023.

\bibitem{codevilla2018end}
Felipe Codevilla, Matthias M{\"u}ller, Antonio L{\'o}pez, Vladlen Koltun, and Alexey Dosovitskiy.
\newblock End-to-end driving via conditional imitation learning.
\newblock In {\em 2018 IEEE international conference on robotics and automation (ICRA)}, pages 4693--4700. IEEE, 2018.

\bibitem{codevilla2019exploring}
Felipe Codevilla, Eder Santana, Antonio~M L{\'o}pez, and Adrien Gaidon.
\newblock Exploring the limitations of behavior cloning for autonomous driving.
\newblock In {\em Proceedings of the IEEE/CVF international conference on computer vision}, pages 9329--9338, 2019.

\bibitem{ettinger2021large}
Scott Ettinger, Shuyang Cheng, Benjamin Caine, Chenxi Liu, Hang Zhao, Sabeek Pradhan, Yuning Chai, Ben Sapp, Charles~R Qi, Yin Zhou, et~al.
\newblock Large scale interactive motion forecasting for autonomous driving: The waymo open motion dataset.
\newblock In {\em Proceedings of the IEEE/CVF International Conference on Computer Vision}, pages 9710--9719, 2021.

\bibitem{freeman2021brax}
C~Daniel Freeman, Erik Frey, Anton Raichuk, Sertan Girgin, Igor Mordatch, and Olivier Bachem.
\newblock Brax--a differentiable physics engine for large scale rigid body simulation.
\newblock {\em arXiv:2106.13281}, 2021.

\bibitem{gillen2022leveraging}
Sean Gillen and Katie Byl.
\newblock Leveraging reward gradients for reinforcement learning in differentiable physics simulations.
\newblock {\em arXiv:2203.02857}, 2022.

\bibitem{gulino2024waymax}
Cole Gulino, Justin Fu, Wenjie Luo, George Tucker, Eli Bronstein, Yiren Lu, Jean Harb, Xinlei Pan, Yan Wang, Xiangyu Chen, et~al.
\newblock Waymax: An accelerated, data-driven simulator for large-scale autonomous driving research.
\newblock {\em Advances in Neural Information Processing Systems}, 36, 2024.

\bibitem{heeg2024learning}
Johannes Heeg, Yunlong Song, and Davide Scaramuzza.
\newblock Learning quadrotor control from visual features using differentiable simulation.
\newblock {\em arXiv preprint arXiv:2410.15979}, 2024.

\bibitem{hochreiter1997long}
Sepp Hochreiter and J{\"u}rgen Schmidhuber.
\newblock Long short-term memory.
\newblock {\em Neural computation}, 9(8):1735--1780, 1997.

\bibitem{holl2020learning}
Philipp Holl, Vladlen Koltun, and Nils Thuerey.
\newblock Learning to control pdes with differentiable physics.
\newblock {\em arXiv:2001.07457}, 2020.

\bibitem{hu2023planning}
Yihan Hu, Jiazhi Yang, Li~Chen, Keyu Li, Chonghao Sima, Xizhou Zhu, Siqi Chai, Senyao Du, Tianwei Lin, Wenhai Wang, et~al.
\newblock Planning-oriented autonomous driving.
\newblock In {\em Proceedings of the IEEE/CVF Conference on Computer Vision and Pattern Recognition}, pages 17853--17862, 2023.

\bibitem{kingma2013auto}
Diederik~P Kingma and Max Welling.
\newblock Auto-encoding variational bayes.
\newblock {\em arXiv:1312.6114}, 2013.

\bibitem{joint_multipath}
Stepan Konev.
\newblock Mpa: Multipath++ based architecture for motion prediction, 2022.

\bibitem{le2022survey}
Luc Le~Mero, Dewei Yi, Mehrdad Dianati, and Alexandros Mouzakitis.
\newblock A survey on imitation learning techniques for end-to-end autonomous vehicles.
\newblock {\em IEEE Transactions on Intelligent Transportation Systems}, 23(9):14128--14147, 2022.

\bibitem{levine2016end}
Sergey Levine, Chelsea Finn, Trevor Darrell, and Pieter Abbeel.
\newblock End-to-end training of deep visuomotor policies.
\newblock {\em Journal of Machine Learning Research}, 17(39):1--40, 2016.

\bibitem{li2024visfly}
Fanxing Li, Fangyu Sun, and Danping Zou.
\newblock Visfly: An efficient and versatile simulator for training vision-based flight.
\newblock {\em arXiv:2407.14783}, 2024.

\bibitem{metz2021gradients}
Luke Metz, C~Daniel Freeman, Samuel~S Schoenholz, and Tal Kachman.
\newblock Gradients are not all you need.
\newblock {\em arXiv:2111.05803}, 2021.

\bibitem{mnih2015human}
Volodymyr Mnih, Koray Kavukcuoglu, David Silver, Andrei~A Rusu, Joel Veness, Marc~G Bellemare, Alex Graves, Martin Riedmiller, Andreas~K Fidjeland, Georg Ostrovski, et~al.
\newblock Human-level control through deep reinforcement learning.
\newblock {\em nature}, 2015.

\bibitem{montali2024waymo}
Nico Montali, John Lambert, Paul Mougin, Alex Kuefler, Nicholas Rhinehart, Michelle Li, Cole Gulino, Tristan Emrich, Zoey Yang, Shimon Whiteson, et~al.
\newblock The waymo open sim agents challenge.
\newblock {\em Advances in Neural Information Processing Systems}, 36, 2024.

\bibitem{mora2021pods}
Miguel Angel~Zamora Mora, Momchil Peychev, Sehoon Ha, Martin Vechev, and Stelian Coros.
\newblock Pods: Policy optimization via differentiable simulation.
\newblock In {\em International Conference on Machine Learning}, pages 7805--7817. PMLR, 2021.

\bibitem{murthy2020gradsim}
J~Krishna Murthy, Miles Macklin, Florian Golemo, Vikram Voleti, Linda Petrini, Martin Weiss, Breandan Considine, J{\'e}r{\^o}me Parent-L{\'e}vesque, Kevin Xie, Kenny Erleben, et~al.
\newblock gradsim: Differentiable simulation for system identification and visuomotor control.
\newblock In {\em International conference on learning representations}, 2020.

\bibitem{nayakanti2023wayformer}
Nigamaa Nayakanti, Rami Al-Rfou, Aurick Zhou, Kratarth Goel, Khaled~S Refaat, and Benjamin Sapp.
\newblock Wayformer: Motion forecasting via simple \& efficient attention networks.
\newblock In {\em 2023 IEEE International Conference on Robotics and Automation (ICRA)}, pages 2980--2987. IEEE, 2023.

\bibitem{schaefer2007recurrent}
Anton~Maximilian Schaefer, Steffen Udluft, and Hans-Georg Zimmermann.
\newblock A recurrent control neural network for data efficient reinforcement learning.
\newblock In {\em 2007 IEEE International Symposium on Approximate Dynamic Programming and Reinforcement Learning}, pages 151--157. IEEE, 2007.

\bibitem{schrittwieser2020mastering}
Julian Schrittwieser, Ioannis Antonoglou, Thomas Hubert, Karen Simonyan, Laurent Sifre, Simon Schmitt, Arthur Guez, Edward Lockhart, Demis Hassabis, Thore Graepel, et~al.
\newblock Mastering atari, go, chess and shogi by planning with a learned model.
\newblock {\em Nature}, 588(7839):604--609, 2020.

\bibitem{schulman2015trust}
John Schulman, Sergey Levine, Pieter Abbeel, Michael Jordan, and Philipp Moritz.
\newblock Trust region policy optimization.
\newblock In {\em International conference on machine learning}, pages 1889--1897. PMLR, 2015.

\bibitem{schulman2017proximal}
John Schulman, Filip Wolski, Prafulla Dhariwal, Alec Radford, and Oleg Klimov.
\newblock Proximal policy optimization algorithms.
\newblock {\em arXiv:1707.06347}, 2017.

\bibitem{shi2022motion}
Shaoshuai Shi, Li~Jiang, Dengxin Dai, and Bernt Schiele.
\newblock Motion transformer with global intention localization and local movement refinement.
\newblock {\em Advances in Neural Information Processing Systems}, 35:6531--6543, 2022.

\bibitem{silver2016mastering}
David Silver, Aja Huang, Chris~J Maddison, Arthur Guez, Laurent Sifre, George Van Den~Driessche, Julian Schrittwieser, Ioannis Antonoglou, Veda Panneershelvam, Marc Lanctot, et~al.
\newblock Mastering the game of go with deep neural networks and tree search.
\newblock {\em nature}, 529(7587):484--489, 2016.

\bibitem{song2024learning}
Yunlong Song, Sangbae Kim, and Davide Scaramuzza.
\newblock Learning quadruped locomotion using differentiable simulation.
\newblock {\em arXiv preprint arXiv:2403.14864}, 2024.

\bibitem{sutskever2014sequence}
Ilya Sutskever, Oriol Vinyals, and Quoc~V Le.
\newblock Sequence to sequence learning with neural networks.
\newblock {\em Advances in neural information processing systems}, 27, 2014.

\bibitem{sutton1999policy}
Richard~S Sutton, David McAllester, Satinder Singh, and Yishay Mansour.
\newblock Policy gradient methods for reinforcement learning with function approximation.
\newblock {\em Advances in neural information processing systems}, 12, 1999.

\bibitem{van2016deep}
Hado Van~Hasselt, Arthur Guez, and David Silver.
\newblock Deep reinforcement learning with double q-learning.
\newblock In {\em Proceedings of the AAAI conference on artificial intelligence}, volume~30, 2016.

\bibitem{vaswani2017attention}
Ashish Vaswani, Noam Shazeer, Niki Parmar, Jakob Uszkoreit, Llion Jones, Aidan~N Gomez, {\L}ukasz Kaiser, and Illia Polosukhin.
\newblock Attention is all you need.
\newblock {\em Advances in neural information processing systems}, 30, 2017.

\bibitem{wang2023multiverse}
Yu~Wang, Tiebiao Zhao, and Fan Yi.
\newblock Multiverse transformer: 1st place solution for waymo open sim agents challenge 2023.
\newblock {\em arXiv:2306.11868}, 2023.

\bibitem{wang2016dueling}
Ziyu Wang, Tom Schaul, Matteo Hessel, Hado Hasselt, Marc Lanctot, and Nando Freitas.
\newblock Dueling network architectures for deep reinforcement learning.
\newblock In {\em International conference on machine learning}, pages 1995--2003. PMLR, 2016.

\bibitem{wiedemann2023training}
Nina Wiedemann, Valentin W{\"u}est, Antonio Loquercio, Matthias M{\"u}ller, Dario Floreano, and Davide Scaramuzza.
\newblock Training efficient controllers via analytic policy gradient.
\newblock In {\em 2023 IEEE International Conference on Robotics and Automation (ICRA)}, pages 1349--1356. IEEE, 2023.

\bibitem{wierstra2010recurrent}
Daan Wierstra, Alexander F{\"o}rster, Jan Peters, and J{\"u}rgen Schmidhuber.
\newblock Recurrent policy gradients.
\newblock {\em Logic Journal of IGPL}, 18(5):620--634, 2010.

\bibitem{yuan2021agentformer}
Ye~Yuan, Xinshuo Weng, Yanglan Ou, and Kris~M Kitani.
\newblock Agentformer: Agent-aware transformers for socio-temporal multi-agent forecasting.
\newblock In {\em Proceedings of the IEEE/CVF International Conference on Computer Vision}, pages 9813--9823, 2021.

\end{thebibliography}
\end{document}